# Scaling Taxonomy-Free Image Labeling with Generative AI

A Multimodal Search Application

Zhenjie Zhang
Pinterest, Inc.
Palo Alto, CA USA
zhenjiezhang@pinterest.com

Sainan Chen
Pinterest, Inc.
Toronto, Canada
sainanchen@pinterest.com

Kshiteesh Hegde
Pinterest, Inc.
Palo Alto, CA USA
khegde@pinterest.com

Mengze (Irene) Li
Pinterest, Inc.
Palo Alto, CA USA
mengzeli@pinterest.com

Pouya Rezazadeh Kalehbasti
Pinterest, Inc.
Palo Alto, CA USA
pouya@pinterest.com

Munachim Ezema
Pinterest, Inc.
Rogers, AR USA
mezema@pinterest.com

Karina Sobhani
Pinterest, Inc.
Palo Alto, CA USA
ksobhani@pinterest.com

## ABSTRACT

Providing real-time multimodal insights into images is challenging because it requires low latency, high description accuracy, and scalable solutions that are both cost-effective and efficient. For example, given a new image of an outfit, we would like to generate insights that the image depicts "earth tones" and "linen pants" and allow users to explore similar ideas along these themes. Traditional strategies to generate such insights would typically involve building taxonomies producing a golden set labeled by humans, training models on these datasets, and deploying these models for online inference services. Such workflows are constrained by fixed taxonomy, high labeling cost, limited labeling coverage in sample space, and human errors and expertise gaps in labelers. In addition, these methods are expensive and cumbersome to iterate over time. Recent trends have started exploring generative AI models for online insights, but this method suffers from long latency and prohibitive costs to support a high volume of requests or large amounts of offline labeling needs.

We present Generative Labeling On Clusters (GLOC) as a solution to provide highly scalable, cost-effective, low latency, and high accuracy online image insight services. GLOC leverages highly optimized image embeddings to form image clusters that are semantically close to each other for the purpose of the type of insights, and summarizes the insights at the cluster level, which effectively generalizes to members in the cluster. At Pinterest, GLOC powers image descriptors and pivot query suggestions for Hybrid Search, which is a multimodal search service enabling users to search with both an image and a text query. Our offline and online A/B tests demonstrate that GLOC significantly outperforms previous methods, delivering improvements in quality and efficiency.

## 1 INTRODUCTION

Providing image insights via short descriptions and query suggestions is a powerful tool in guiding a user's journey and fulfilling user intent on visual platforms. At Pinterest, we actively develop more efficient image insight extraction solutions with high accuracy and low cost. Hybrid Search is a new service at Pinterest that enables users to search with a combination of an image and an accompanying text query. We provide query suggestions for Hybrid Search for easier interaction with this tool. These query suggestions are in two forms of insights: descriptors and pivots. Descriptors are concise, open-ended phrases that capture key aspects of an image (fashion style, color palettes, room aesthetic, etc.) that help users to better articulate what aspect of an image may be resonating with them. Pivots are short, actionable phrases that enable users to modify images as shown in Figure 1. These modifications can be subtle, like adjusting details (color, style, etc.), or more substantial and involve significant changes of certain aspects of the image (e.g., change this blue floral dress to knee length, instead of ankle length).

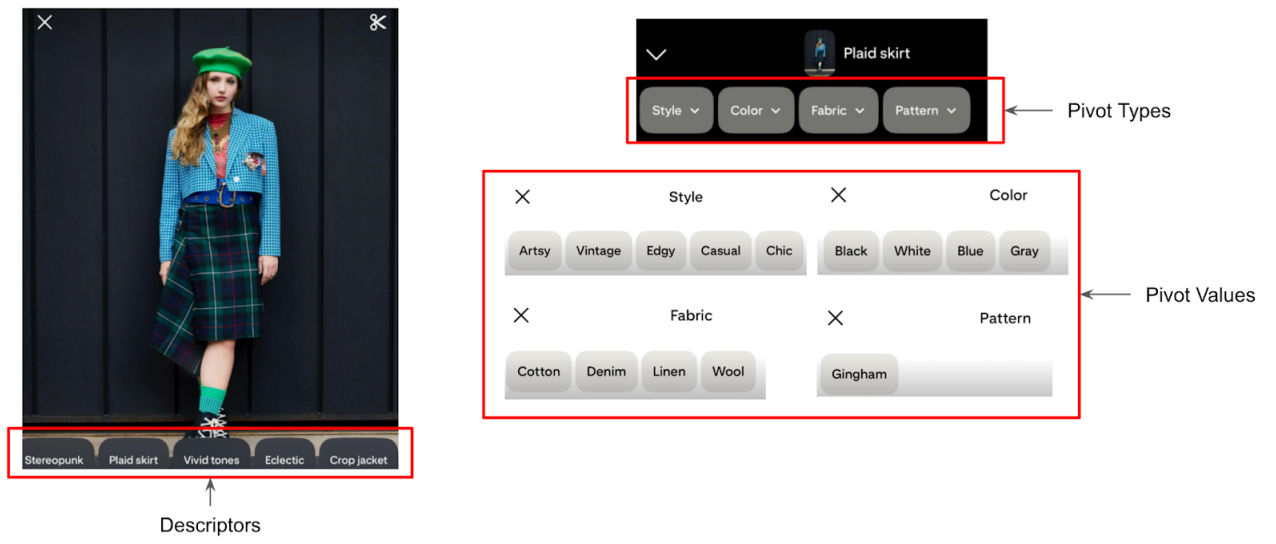

**Figure 1: Illustrations of how descriptors and pivots appear on an image within Pinterest's Visual Search. Descriptors are concise phrases describing aspects of the image, and pivots are short, actionable phrases for modifying images.**

Coming up with relevant descriptors for an image is challenging, as these descriptors need to cover a vast range of possible scenes, which involves a broad and deep understanding of the real world in order to make the suggestions interesting and helpful. A predictive model based on a taxonomy [3, 6, 7] tends to fall short due to the limitation imposed by the fixed taxonomy. A visual language model [2, 4, 8] provides richer insights, yet the inference cost on billions of images is prohibitively high. In addition, these models are often too slow for online services, and are very time-consuming even for offline generation on large scale image datasets [9, 11].

Pivot suggestions have similar challenges, with the additional complication that the applicable modifications are more subjective and are context dependent. Suggesting "change to gray color" is perfectly acceptable on a blue shirt but is inappropriate for a blue gemstone. Training a predictive model faces additional complexity as there are more than a few appropriate answers to each type of pivot.

To address these challenges, we developed Generative Labeling On Clusters (GLOC), a method that effectively scales up the efficiency of visual language model generated insights for both the descriptors and pivots. GLOC leverages large-scale clustering of visual embeddings, ensuring high semantic relevance within each cluster. We discuss the key considerations and refinements involved in building this system and demonstrate the effectiveness of the methodology with significant offline and online gains.

## 2 METHODOLOGY

### 2.1 Summary of the GLOC System Components

The GLOC System comprises an offline workflow and an online serving architecture as shown in Figure 2. The offline workflow performs large scale clustering [5] to generate fine granularity image clusters, propose representative images for each cluster, use visual language models to produce insights with representative images, and upload these insights to the store. The online architecture includes a pytorch model or an ANN cluster that finds cluster membership with very short latency and looks up the insights in a kv-store.

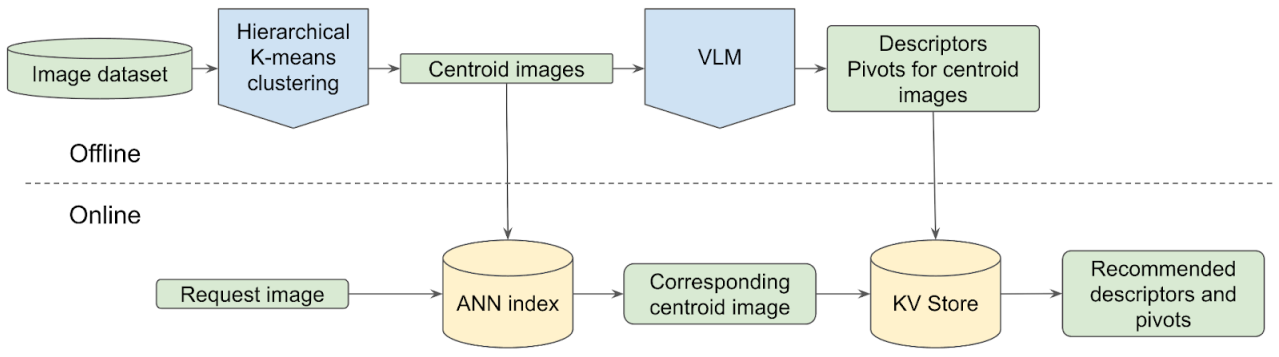

**Figure 2: Overview of the Generative Labeling On Clusters (GLOC) system. The offline stage includes large-scale image clustering, representative images selection, and insights insight labeling with VLM. The online serving architecture includes efficient cluster membership lookup and retrieval of stored insights for real-time image analysis.**

To make this system work effectively, there are a number of considerations to be taken into account.

- Efficient clustering method
- Choice of image embedding
- Clustering density
- Online service efficacy

### 2.2 Details of the GLOC components

*2.2.1 Hierarchical K-means Clustering.* To scale up clustering to handle billions of images, we developed hierarchical K-means clustering. In this procedure, we first use k-means to split the dataset into a small number of partitions, each being a coarse level cluster. Each of these partitions are of manageable size so that they can fit in the GPU memory of an AWS g5.12xlarge instance. A second stage of clustering is conducted on each of the partitions to find the final fine granular clusters.

In order to represent each cluster with actual images, we perform nearest neighbor search within all images of each partition with the cluster centroid embeddings and take the top images to represent the cluster. These images are used for VLM labeling to capture the image insights of the cluster.

At the final stage, we combine all of the centroid embeddings from all partitions into one index for cluster membership search.

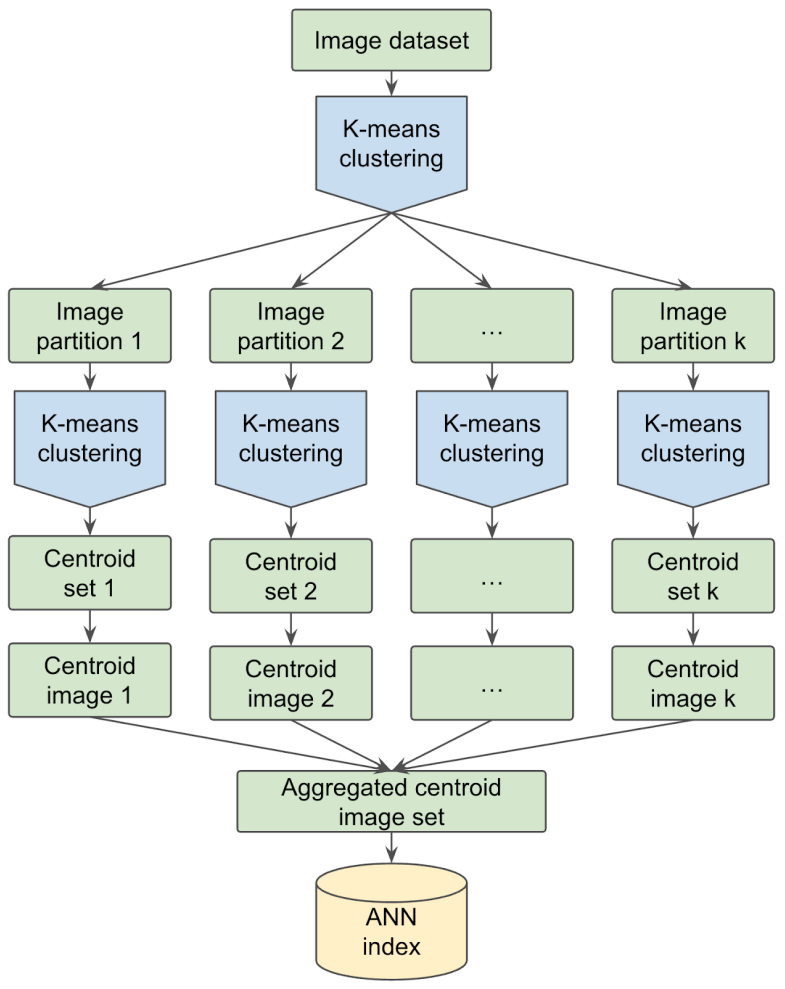

**Figure 3: Hierarchical K-means clustering framework (assuming taking 1 centroid image for each cluster).**

The time complexity of the classic K-means algorithm is $O(NTK)$. With $N$ being the total data points, $T$ being the number of iterations and $K$ being the number of clusters. When we fix the ratio between $N$ and $K$, the time complexity becomes quadratic in $N$: $O(N^2T/C)$ with the constant $C$ being the average cluster size: $N/K$.

GLOC allows us to avoid this quadratic time complexity, thus avoiding the need for an expensive distributed GPU cluster with the growth of $N$. With GLOC, the time complexity is linear in $N$ as $O(NST/C)$, with $S$ being the partition size, which we keep as a constant.

The quality of clustering should be the same to classic k-means at the interior of each partition, but can drop slightly at the borders between clusters, as the average distance between member and centroid can double at these regions. We take this as an acceptable trade-off**.**

*2.2.2 Embedding*. GLOC relies on high quality image embeddings to gain semantic generalization within clusters. It is critical to choose or develop an embedding that fits the specific type of insights that the online system intends to provide. Meanwhile, this embedding needs to be of manageable size, so as to perform clustering efficiently. For Hybrid Search Pivots and Descriptors, we use Unified Visual Embedding developed at Pinterest [10], for the prowess of this embedding in modeling highly detailed similarity in visual perception. We found that using embeddings which are not specialized strictly for semantic relevance, but also include engagement optimization [1] lead to inferior results. Depending on the types of insights, GLOC can use specialized embeddings. For example, an embedding specifically encoding colors can achieve superior results on color insights compared to other embeddings.

*2.2.3 Clustering density***.** The quality of results also relies on the density of clusters produced to capture high granularity insights. We compared different clustering density and achieved an optimal balance between quality and cost with an average cluster size of 2000.

2.2.4 Online service. There are two ways to look up cluster membership online. For a small number of centroids (~ 1M), we use a single layer neural network model to encode the centroid embeddings. A membership lookup is equivalent to a neural network inference. For a large number of centroids, we use an ANN index for efficient membership lookup. Once the membership is identified, the cluster identify is used to look up in the KV store for corresponding descriptors and pivots.

# 3 RESULTS

## 3.1 Cost saving

We use a clustering size averaging 2000. This enabled us to reduce the label generation cost to 0.05%. Use cases of different complexity may need adjustments in cluster size to trade-off between cost and quality. Our results demonstrate the methodology is able to keep the cost of generative AI labeling low.

## 3.2 Offline Evaluation

We've productionized GLOC system for multimodal image descriptors and pivot suggestions at Pinterest. To assess the quality of these text insights, we conducted human evaluation on the relevance of text output to the images. For pivots, we have four levels of applicability: very close to the image (score 4), very reasonable pivot to apply to the image (score 3), relevant pivot to apply to the image (score 2), irrelevant pivot for the image (score 1). GLOC achieved an average score of 3.5 out of 4. For descriptors, there are three relevance levels: highly relevant (score 3), partially relevant (score 2), not relevant (score 1). The GLOC system achieved an average score of 2.51 out of 3, with a low irrelevant rate of 4%.

## 3.3 Online A/B Experiment

We ran an A/B experiment where users are randomly selected to either not receive descriptor and pivot experience or get recommendations of descriptors and pivots on eligible images. When a user clicks on a descriptor or a pivot, we generate a multimodal feed that is relevant to both the text and the image (how this feed is generated is out of the scope of this paper).

For the pivot experiment, there are three groups.

- Control: users see no pivot.
- Baseline Treatment (BT): users see pivots, which are assigned by text based GPT with a text summary of the image.
- GLOC: users see pivots assigned by GLOC method.

The GLOC group outperformed both the control and the BT group in terms of both pivot CTR and feed engagement rate. We attribute the improvement in feed engagement (GLOC vs BT) to the improvements of pivot relevance leading to the feasibility of achieving higher relevance in feed.

| GLOC Performance[1] | Pivot CTR | Pivot Feed engagement | Descriptor feed engagement |
|---|---|---|---|
| Compared to control | NA | +146% | +79% |
| Compared to Baseline Treatment (BT) | +69.5% | +100% | NA |

**Table 1: Online A/B test results for pivot and descriptor suggestions, demonstrating significant improvements in CTR and Feed Engagement in the GLOC treatment groups compared to the control and baseline treatment. Pivot CTR was not compared to control, as there are no pivots for clicking in the control group. Descriptor feed engagement was not compared to the BT group, as there is no naive candidate to implement this baseline, which is taxonomy-free.**

For the descriptor experiment, there are two groups:

- Control: users see no descriptors.
- GLOC: users see descriptors assigned by GLOC method.

We observed similar improvements in Feed engagement of +79% compared to control (we did not have a BT group for this experiment).

## ACKNOWLEDGMENTS

We would like to thank our colleagues at Pinterest for contributing to the work in this paper: Shirley Du, Kayoung Lee, Nikita Hudson, Ocie Henderson, Eduardo Lopez, Alejandra Estefania Garza Martinez, Helen Xu, Cindy Zhang, Anatolii Shevchenko, Rex Wu, Florian Marcu, Sai Xiao and Raja Raina for their support and insights.

[1] Pinterest Internal Data. A/B experiment in US and Canada, 2025.